\documentclass{article}

\PassOptionsToPackage{numbers, compress}{natbib}
 \usepackage[preprint]{neurips_2026}

\usepackage[utf8]{inputenc} 
\usepackage[T1]{fontenc}    
\usepackage{hyperref}       
\usepackage{url}            
\usepackage{booktabs}       
\usepackage{amsfonts}       
\usepackage{nicefrac}       
\usepackage{microtype}      
\usepackage{xcolor}         
\usepackage{graphicx} 
\usepackage{hyphenat}
\usepackage{amsmath}
\usepackage{multirow}
\usepackage{arydshln}
\usepackage{subcaption}
\usepackage{booktabs}
\usepackage{wrapfig}
\usepackage{makecell}
\usepackage{pifont}
\newcommand{\ours}{Point4D }

\usepackage{algorithm}
\usepackage{algpseudocode}
\usepackage{amsmath}

\usepackage{bbm}
 
\newcommand{\xmark}{\ding{55}}
\newcommand{\cmark}{\ding{51}}

\renewcommand{\paragraph}[1]{\vspace{0.25em}\noindent\textbf{#1}\quad}

\usepackage[table]{xcolor}
\definecolor{gold}{RGB}{255,223,0}
\definecolor{silver}{RGB}{255,165,0}  
\definecolor{bronze}{RGB}{255,255,0}  

\definecolor{rank1}{RGB}{255,100,100}   
\definecolor{rank2}{RGB}{255,165,80}    
\definecolor{rank3}{RGB}{255,235,80}    

\title{Point4D: Long-range 4D Motion Reconstruction 
}

\author{%
  Minsik Jeon \quad Jay Karhade \quad Deva Ramanan$^\dagger$ \quad Shubham Tulsiani$^\dagger$ \\
  Carnegie Mellon University \\
  \textit{Project Page:} \href{https://point-4d.github.io}{https://point-4d.github.io}
}

\begin{document}

\maketitle
\begingroup
\renewcommand\thefootnote{}\footnotetext{$^\dagger$Equal advising.}%
\addtocounter{footnote}{-1}%
\endgroup

\begin{figure}[h]
  \centering
  \includegraphics[width=1.0\linewidth]{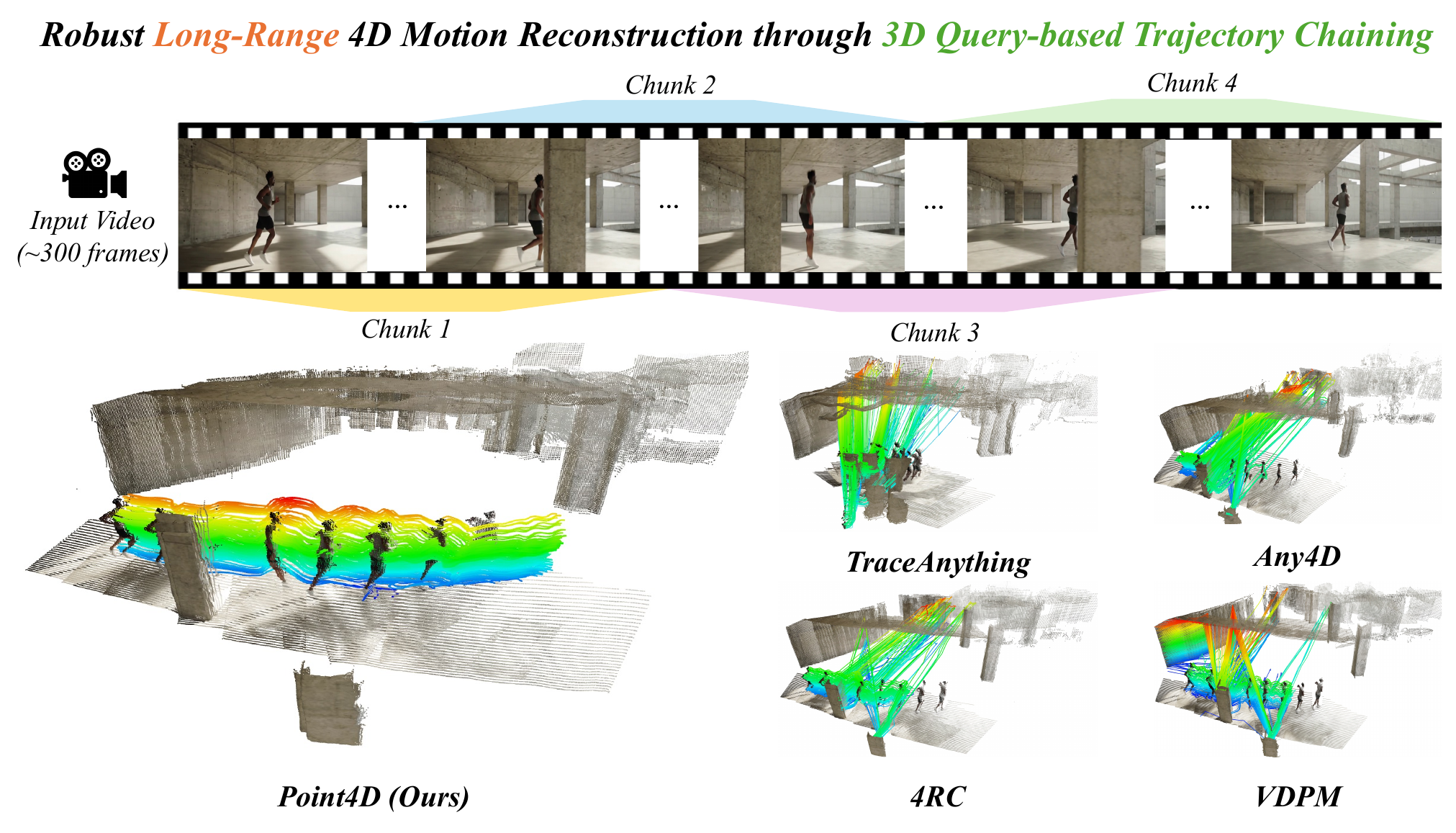}
  \caption{
\textbf{\ours{} enables long-range 4D reconstruction via autoregressively chaining motion reconstruction across overlapping chunks}. Existing 4D reconstruction methods (e.g., TraceAnything~\cite{traceany}, Any4D~\cite{any4d}, 4RC~\cite{4rc}, VDPM~\cite{vdpm}) predict 4D motion for query 2D pixels, and cannot be easily chained under occlusion as the tracked points may not be visible in overlapping frames. \ours decodes motion for query 3D points instead of 2D pixels, allowing direct chaining for long-range motion reconstruction under occlusions.
}
  \label{fig:teaser}
\end{figure}

\begin{abstract}
We introduce Point4D, a feed-forward model for 4D reconstruction of long-range video sequences. Point4D is able to reliably infer dense per-point 3D trajectories across multi-hundred-frame videos, unlike existing 4D methods that are limited to short input windows of at most a few dozen frames. A key innovation that enables this is our flexible 3D query-based motion decoder that decouples trajectory prediction from image-plane visibility. The predicted 3D endpoints are then directly re-queried in the next chunk without re-projection or matching. Furthermore, we show that extracting and reusing a visual descriptor from an arbitrary frame where the point is visible leads to better performance than relying solely on the source patch. Overall, Point4D achieves state-of-the-art performance across diverse long-video tracking benchmarks spanning over 200 frames and largely outperforms previous feed-forward 4D method. Project page: \href{https://point-4d.github.io}{point-4d.github.io}
\end{abstract}

\section{Introduction}

Consider the runner in Fig.~\ref{fig:teaser}: a person jogging down a long corridor, vanishing behind structural columns and reappearing moments later. We humans can effortlessly follow the runner's continuing trajectory throughout the clip, understanding their long-range motion despite occlusions, and even reasoning about where they may be when not directly visible. In this work, we seek to build a computational system that can similarly perform \emph{long-range 4D motion reconstruction} from a monocular video --- recovering per-point 3D trajectories across hundreds of frames.

While initial approaches~\cite{wang2023omnimotion, shapeofmotion, tapip3d, spatrack} for such `4D reconstruction' leveraged slow and expensive iterative optimization, there has been a marked shift towards feed-forward methods~\cite{any4d, d4rt, 4rc, flow4r, traceany, vdpm} which, following the successes in feed-forward 3D reconstruction~\cite{dust3r, vggt}, have extended such multi-view models to additionally perform motion prediction. Specifically, by adding scene flow prediction `heads' to multi-view models, these approaches allow decoding each pixel's 3D position at any queried timestep, thereby inferring the 4D motion of the scene in an efficient feed-forward manner. While these methods deliver impressive results, they are designed to operate over short input windows of a handful of frames. Long videos that span hundreds or more frames expose two challenges that short windows do not face. \textbf{First}, jointly processing the full frame set is computationally infeasible for the underlying transformer encoders, and \textbf{second}, the points to be tracked routinely become occluded or leave the field of view between observations. 

\label{sec:Introduction}
\begin{wrapfigure}{r}{0.5\linewidth}

  \centering
  \includegraphics[width=\linewidth]{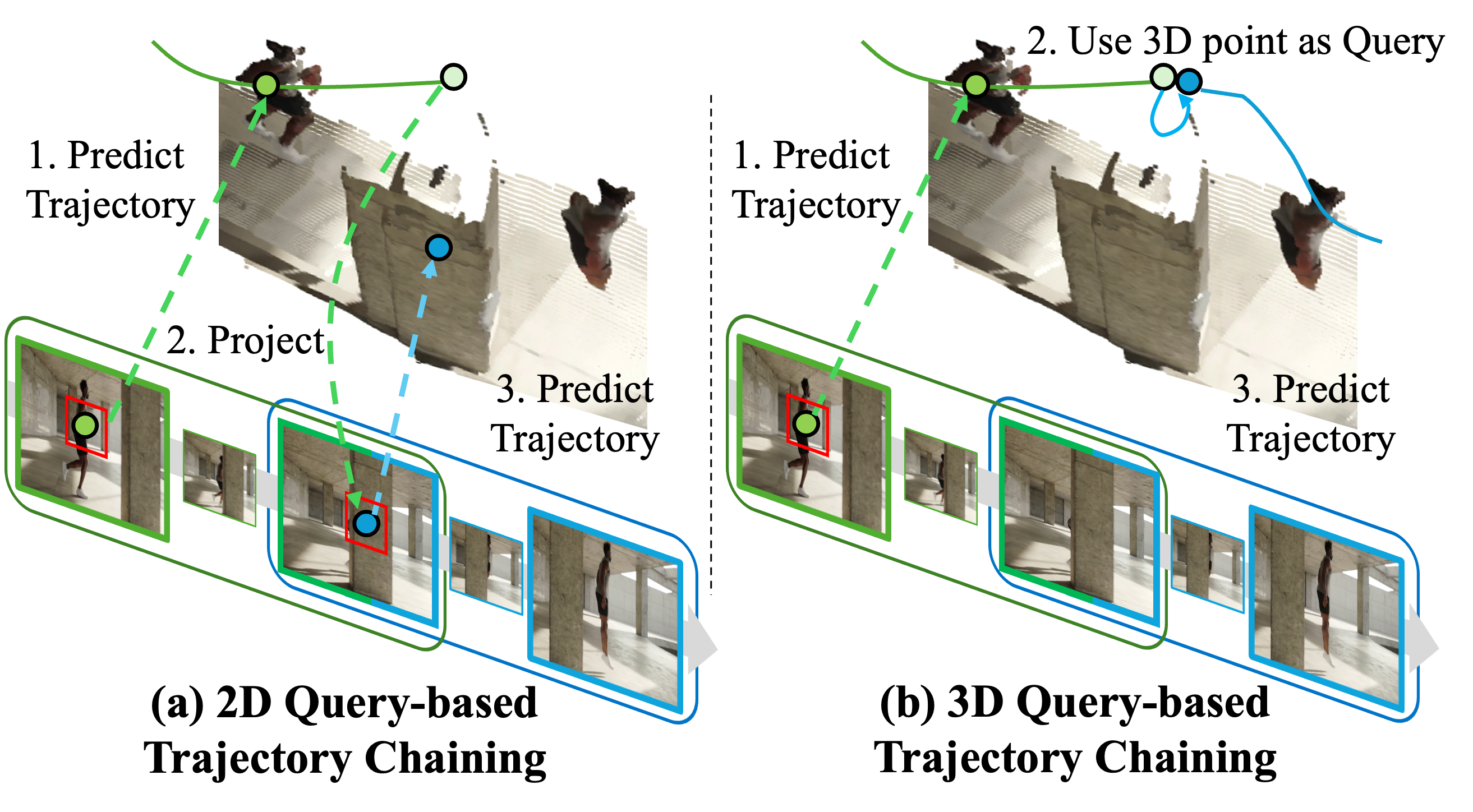}
  \caption{\textbf{\ours uses 3D query representations for chaining motion prediction.} A 2D-query-based motion decoder requires image reprojection and is not robust to chaining under occlusions. Our insight is to leverage 3D queries, allowing direct chaining without reprojection.}
  \label{fig:approach}
\end{wrapfigure}

The first challenge is not unique to long-range 4D reconstruction. Indeed, approaches for long-range static reconstruction are similarly bottlenecked by computational complexity, and a common solution is to \emph{chain} short-range predictions across overlapping chunks~\cite{vggtlong}. Whereas chaining static reconstruction merely requires aligning the coordinate systems from independent per-chunk reconstructions, \emph{chaining for 4D reconstruction requires autoregressive motion prediction} -- the motion for a point tracked in an earlier chunk needs to be queried again from a later chunk to continue the point track. Existing 4D methods that compute 3D track predictions for query \emph{pixels} are not well suited for such chaining, either requiring reprojection from 3D to 2D for re-querying or a computationally expensive `matching' of independently predicted 4D trajectories. However, these strategies break down when a point is occluded or outside the field of view at the chunk boundary which are the very situations that are routine in long videos. Our key insight is to \textbf{query points, not pixels} --  a motion reconstruction decoder with 3D points as queries allows the predicted endpoint of one chunk to be re-queried in the next directly, with no reprojection or matching (see Fig. \ref{fig:approach}). More generally, 3D queries also encourage the decoder to focus on motion reconstruction instead of geometry estimation (which is already captured in the input) and can also help resolve ambiguous 2D pixel queries e.g., on object boundaries where it may be unclear if the query is a foreground or background pixel.

We operationalize this insight in \ours, a feed-forward model that decodes 3D motion from a 3D-coordinate query. A shared ViT encoder produces a scene representation along with per-frame depth and camera poses; a lightweight cross-attention decoder takes a 3D query, target time and camera indices, and a visual descriptor extracted once from any prior frame where the point is visible; it then predicts the queried point's 3D position at the target time. To process long videos, we partition the video into overlapping chunks, autoregressively re-querying each predicted 3D endpoint in the next chunk based on the decoded trajectory from the previous one, while reusing the same visual descriptor throughout. We validate our approach across multiple datasets and show that \ours{} substantially improves over the prior state of the art on long-video 4D tracking while also improving over its 2D-query counterpart for short-range motion reconstruction.

\pagebreak
In summary, our contributions are:
\begin{itemize}
    \item We build a feed-forward pipeline for \emph{long-range} 4D motion reconstruction, recovering per-point motion across hundreds of frames via trajectory chaining across chunks.
    \item We introduce a 3D-query-based motion decoding that  decouples trajectory prediction from image-plane visibility. This also allows predicted 3D endpoints to be propagated directly across chunks.
    \item We achieve state-of-the-art long-video 4D tracking, outperforming both chaining-based feed-forward 4D methods and 3D point trackers.
\end{itemize}

\section{Related Work}
\label{sec:Related_work}

{\bf 3D Tracking.} To model scene motion, point tracking and optical flow~\cite{raft} methods estimate pixel-level correspondences across frames. Subsequent work extended tracking to longer temporal ranges through correlation-based matching and iterative updates~\cite{tapvid, tapir, cotracker3}, yet these methods operate purely in 2D. The TAPVid-3D benchmark~\cite{tapvid3d} motivated lifting tracks into 3D: DELTA~\cite{delta} and SpatialTracker~\cite{spatrack} combine 2D trackers with monocular depth to track points in camera coordinates. More recently, TAPIP3D~\cite{tapip3d} lifts video features into a camera-stabilized 3D point cloud and iteratively refines trajectories in 3D space, while SpatialTrackerV2~\cite{spatrackv2} learns geometry and point motion jointly in an end-to-end architecture, eliminating separate depth estimation. However, these methods only track sparse query points, often rely on off-the-shelf depth modules, and require iterative refinement that limits speed. \ours~can predict dense 3D trajectories in a single forward pass via query-based decoding, while remaining competitive with iterative trackers on sparse benchmarks.

{\bf 4D Reconstruction of Dynamic Scenes.}
4D reconstruction aims to recover both the 3D structure of a scene and how it changes over time. Early methods rely on per-scene optimization~\cite{shapeofmotion}, achieving high fidelity but at a prohibitive cost. Recent feed-forward approaches extend 3D reconstruction architectures to the temporal dimension, predicting dense geometry and motion jointly from a short window of frames~\cite{startrack, traceany, any4d, vdpm, 4rc}. Although these methods differ in motion representation -- scene flow from a canonical view~\cite{any4d}, continuous trajectory fields~\cite{traceany}, or dense point maps at queried timesteps~\cite{4rc, vdpm}, they all decode motion for every pixel through DPT-style heads or similar dense decoders. D4RT~\cite{d4rt} instead introduces a query-based 4D decoder that predicts the 3D position of a 2D query point at a target timestep and coordinate frame. In all cases, however, queries live on the image plane, so extending a track beyond the window requires reprojecting predictions back to pixels, which fails under occlusion. Together with the memory cost of joint multi-frame encoding, this confines prior work to a few dozen frames. We extend D4RT's query-based decoding from 2D to 3D queries, which decouples tracking from visibility and enables longer video 4D reconstruction to over hundreds of frames through direct re-querying of predicted 3D positions.

{\bf Long-range 3D reconstruction.}
Feed-forward 3D reconstruction methods ~\cite{dust3r, vggt, da3, keetha2026mapanything} have shown impressive results in multi-view 3D reconstruction. However, these approaches only work for a limited number of frames, due to high memory requirements, and quickly run out of memory for longer sequence inputs. CUT3R~\cite{wang2025continuous}, InfiniteVGGT ~\cite{yuan2026infinitevggt}, StreamingVGGT ~\cite{zhuo2025streaming} proposed memory mechanisms for long-sequence reconstructions, while VGGT-Long ~\cite{vggtlong} proposes a chunking and loop closure mechanism. More recently, Loger~\cite{zhang2026loger}, TTT3R~\cite{ttt3r} and Zipmap~\cite{zipmap} integrate test-time training mechanisms. These works demonstrate that 3D reconstruction models trained for small inputs can be effectively chained to handle longer sequences. \ours{} extends this insight to the 4D setting: rather than simply chaining static geometry, it uses 3D point queries to chain motion across chunks, recovering dense scene flow and long-range 3D tracks in a scalable, feed-forward manner.
\section{Method}
\label{sec:Method}

\begin{figure}[t]
  \centering
  \includegraphics[width=1.0\linewidth]{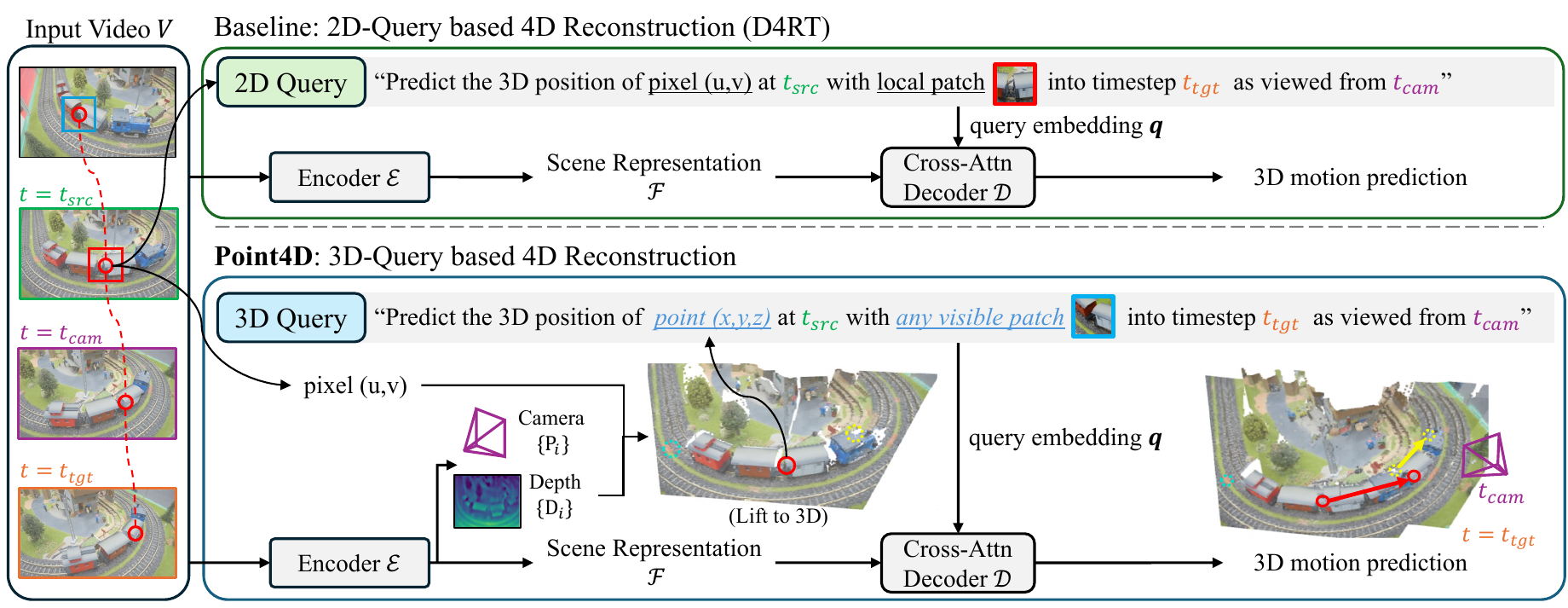}
  \caption{\textbf{Overview of \ours.} Unlike 2D-query methods, which bind a query to a source pixel and its patch, \ours queries a 3D point with a descriptor drawn from any frame in which it is visible. Given a monocular video $V$, an encoder $\mathcal{E}$ with self-attention layers produces a scene representation $\mathcal{F}$ together with per-frame depth maps and camera poses. A 3D query $\mathbf{p}=(x,y,z)$ is constructed as the point's position at time $t_\text{src}$, expressed in the camera coordinates of frame $t_\text{src}$, and its visual patch is extracted from any timestep where the point is visible. The resulting query embedding $\mathbf{q}$ cross-attends to $\mathcal{F}$ to predict the point's 3D position at the target timestep $t_\text{tgt}$. Together, the 3D query and its visibility-agnostic descriptor decouple a query point from any single frame's image plane, so that occluded and out-of-frame points remain valid queries.}
  \label{fig:model}
\vspace{-0.1in}
\end{figure}

Our goal is to recover dense 3D trajectories from long monocular video sequences. To achieve this, we propose \textbf{\ours}, a feed-forward model that uses \textit{3D coordinate queries} as input instead of inferring motion from 2D pixels. By querying directly in 3D space, we decouple trajectory prediction from image-plane visibility. Building upon established visual geometry prediction models~\cite{vggt,da3,4rc} and D4RT~\cite{d4rt}, we first encode the video into a feature representation while predicting depth and camera poses to support query decoding (Sec.~\ref{sec:3.1.prelims}). We then reformulate the query mechanism to use 3D points $(x, y, z)$ in the source frame, paired with a visual descriptor extracted from any timestep at which the point is visible (Sec.~\ref{sec:3.2.3dquery}). This formulation allows us to predict a point's 3D position at any target time regardless of its image-plane visibility. Finally, we present a simple chaining framework for long-range 3D trajectory inference (Sec.~\ref{sec:3.3.method_chaining}). Since 3D queries remain well-defined under occlusion, we can propagate trajectories across overlapping video chunks by re-querying predicted 3D endpoints, remaining valid through occlusions and field-of-view exits. An overview of our method is shown in Fig.~\ref{fig:model}.

\subsection{Preliminaries}
\label{sec:3.1.prelims}
\paragraph{Feed-forward visual geometry prediction.}
Recent feed-forward methods~\cite{da3, vggt, monst3r, dust3r, flow3r} process multi-view images, or video $V$  through a ViT backbone $\mathcal{E}$ with alternating frame-wise and global self-attention layers, producing patch tokens $\mathbf{Z}_i$ and camera tokens $\mathbf{c}_i$ for each frame $i \in \{1,\dots,T\}$. Dedicated heads decode camera poses $\mathbf{P}_i$ from $\mathbf{c}_i$ and depth maps $\mathbf{D}_i$ from $\mathbf{Z}_i$ via a DPT decoder~\cite{dpt}. To handle dynamic scenes, recent work augments this representation with a learnable time token $\mathbf{t}_i$ per frame~\cite{4rc}, initialized from a sinusoidal encoding of the normalized timestep in $[0,1]$ and refined through the same attention layers. This yields a scene representation $\mathcal{F}=\{\mathbf{Z}_i\}_{i=1}^T$ along with per-frame tokens and predictions:
\begin{equation}
    \mathcal{F}, \{\mathbf{c}_i,\mathbf{t}_i,\mathbf{P}_i,\mathbf{D}_i\}_{i=1}^T=\mathcal{E}(V).
\end{equation}

\paragraph{Query-based 4D decoding.}
While most 4D reconstruction methods utilize DPT-style heads for dense geometry, D4RT introduces on-demand query-based decoding. This mechanism defines a query as $(u, v, t_\text{src}, t_\text{tgt}, t_\text{cam}, S)$, where $(u, v)$ represents a 2D pixel in frame $t_\text{src}$ and $S$ provides visual context via an RGB patch around $(u, v)$. The decoder $\mathcal{D}$ then predicts the 3D position $\hat{\textbf{p}}$ of the corresponding point pixel at a target timestep $t_\text{tgt}$, expressed in camera $t_\text{cam}$'s coordinate system. Each query is mapped to an embedding $\mathbf{q}$ and independently cross-attends to $\mathcal{F}$ to produce a predicted point:
\begin{equation}
    \hat{\mathbf{p}} = \mathcal{D}(\mathbf{q}, \mathcal{F}) \in \mathbb{R}^3.
\label{eq:decode}
\end{equation}
This formulation enables flexible decoding of arbitrary pixel sets across varying timesteps and coordinate frames. 
However, like prior 4D methods, D4RT relies on 2D pixel coordinates, requiring the queried point to be visible in the source frame for initialization.

\subsection{3D Query-based Decoder}
\label{sec:3.2.3dquery}

Existing 4D reconstruction methods infer motion via predicting scene flow for 2D pixels and require the tracked 3D point to be visible in the source frame. This breaks down when the point is occluded or outside the field of view, and is especially problematic across video chunks where the predicted 3D position at the end of one chunk may have no visible pixel to re-query in the next. We address this with two key design choices. \textbf{First}, we replace 2D pixels with \textit{3D points} as queries, which decouples point identity from image-plane visibility, and allows the decoder to focus on predicting motion rather than jointly recovering geometry. \textbf{Second}, we supply appearance context via a local image patch drawn from \textit{any} frame where the point is visible, rather than tying visual context to the query frame. This enables cross-chunk trajectory chaining as the patch can be sourced from a different chunk entirely, even when the point is occluded or out of view in the current one.

\paragraph{3D query construction.}
To specify a query, we first select a pixel $(u, v)$ that is visible in some frame $t_\text{ref}$ and extract a local image patch $S$ around it as a visual descriptor. We then obtain the query coordinate $\mathbf{p} = (x, y, z)$ by taking this point's 3D position at a source time $t_\text{src}$, expressed in frame $t_\text{src}$'s camera coordinate system. When $t_\text{ref} \neq t_\text{src}$, the point may have moved and may be occluded or outside the field of view at $t_\text{src}$. While its corresponding 2D pixel is then ill-defined, the 3D coordinate remains valid since it does not depend on the pixel projection. The full query is $(\mathbf{p},\, t_\text{src},\, t_\text{tgt},\, t_\text{cam},\,S)$: \textit{where is the 3D point at position $\mathbf{p}$ (in frame $t_\text{src}$'s coordinate frame) with visual descriptor $S$ at timestep $t_\text{tgt}$, expressed in camera $t_\text{cam}$'s coordinate frame?}

\paragraph{Query embedding and decoding.}
To build the query embedding $\mathbf{q}$, the 3D spatial coordinates $\mathbf{p}$ are first encoded with sinusoidal positional encoding~\cite{attention}. The temporal and camera indices are encoded using corresponding tokens from the encoder: $t_\text{src}$ and $t_\text{cam}$ use camera tokens $\mathbf{c}_{t_\text{src}}$ and $\mathbf{c}_{t_\text{cam}}$, while $t_\text{tgt}$ uses time token $\mathbf{t}_{t_\text{tgt}}$. These encodings are summed with an embedding of $S$ to form the query embedding $\mathbf{q}$. The decoder then follows D4RT~(Eq.~\ref{eq:decode}), passing $\mathbf{q}$ through cross-attention layers attending to $\mathcal{F}$ and obtain predicted 3D position $\hat{\mathbf{p}}$. It uses no self-attention between queries, so each query is decoded independently, enabling flexible batching of arbitrary query sets at inference.

\paragraph{Loss.}
The primary loss $\mathcal{L}_\text{point}$ is an L1 loss on the predicted 3D position, where both prediction and target are passed through a signed log-transform $\phi(x) = \operatorname{sign}(x)\log(1+|x|)$ to dampen the influence of far-away points. The confidence loss $\mathcal{L}_\text{conf}$ modulates $\mathcal{L}_\text{point}$ by a per-query confidence score, so that uncertain predictions are penalized less heavily. The auxiliary reprojection loss $\mathcal{L}_\text{2d}$ is an L1 loss on the predicted 2D projection of $\hat{\mathbf{p}}$ into frame $t_\text{cam}$, enforcing consistency between the predicted 3D position and camera geometry. The auxiliary visibility loss $\mathcal{L}_\text{vis}$ is a binary cross-entropy loss on a per-query logit predicting whether the point is visible at $t_\text{tgt}$. The total loss is:
\begin{equation}
    \mathcal{L} = \mathcal{L}_\text{point} + \mathcal{L}_\text{conf} + \mathcal{L}_\text{2d} + \mathcal{L}_\text{vis}.
\end{equation}

\subsection{Trajectory Chaining}
\label{sec:3.3.method_chaining}

Long videos cannot be processed in a single forward pass due to memory constraints. We therefore partition the video into short, overlapping chunks and encode each chunk independently. Producing \emph{coherent long-range trajectories} from these chunk-level predictions requires (1) aligning each chunk's independent 3D coordinate frame into a single global frame, and (2) propagating point identity across chunk boundaries so that per-chunk predictions form a single continuous trajectory.

\begin{algorithm}
\caption{Trajectory Chaining for Long Video}
\label{alg:traj_chain}
\begin{algorithmic}[1]
\Require Video $V$ of $T$ frames; query pixels $\{(u_i, v_i)\}$ visible in frame 0
\Ensure 4D trajectories $\{P_i(t)\}_{t=0}^{T-1}$ in a global coordinate frame

\State Partition $V$ into $K$ overlapping chunks $\{C_k\}_{k=0}^{K-1}$, where $C_k$ starts at global time $t_k$
\State $\mathbf{p}_i \leftarrow \textsc{Unproject}(u_i,\, v_i,\, \mathbf{D}_0)$ \quad $\forall i$ \Comment{lift pixel query to 3D at $t{=}0$}
\State $S_i \leftarrow \textsc{ExtractPatch}(\mathrm{frame}_0,\, u_i,\, v_i)$ \quad $\forall i$ \Comment{visual descriptor, extracted once}

\For{$k = 0, \ldots, K-1$}
    \State $F_k \leftarrow \textsc{Encode}(C_k)$ \Comment{4D scene representation for chunk $k$}
    \State $\pi_t \leftarrow (t_{src}=0,\; t_{tgt}=t - t_k,\; t_{cam}=0)$ \Comment{$(t_{src}, t_{tgt}, t_{cam})$ local to $C_k$}
    \State $P_i(t) \leftarrow \mathcal{D}((\mathbf{p}_i,\; \pi_t,\; S_i),\; F_k)$ \quad $\forall i,\; \forall t \in C_k$ \Comment{decode 3D position at each frame}
    \If{$k < K-1$}
        \State $(c,\mathbf{R},\mathbf{t}) \leftarrow \textsc{EstimateSim3}(C_k \cap C_{k+1})$ \Comment{align chunks via overlap depth}
        \State $\mathbf{p}_i \leftarrow c\,\mathbf{R}\,P_i(t_{k+1}) + \mathbf{t}$ \quad $\forall i$ \Comment{propagate 3D query into $C_{k+1}$'s frame; reuse $S_i$}
    \EndIf
\EndFor
\State \Return $\{P_i(t)\}_{t=0}^{T-1}$ transformed to global frame-0 coordinates
\end{algorithmic}
\end{algorithm}

\paragraph{Alignment of chunk-level 4D reconstruction.} Since each chunk's geometry is predicted up to an unknown scale and its own coordinate frame, we align adjacent chunks via a Sim(3) transformation~\cite{umeyama} estimated from dense depth predictions on the shared overlap frames. Composing these pairwise Sim(3) transforms places every chunk's predictions into a single global coordinate frame. While this procedure suffices for static 3D reconstruction, we also need to propagate the \textit{temporal point identity} across chunks for 4D scene reconstruction.

\paragraph{Chaining trajectories with 3D queries.} Our 3D query formulation reduces this correspondence problem to a single coordinate transform. Within each chunk, we decode every query point's trajectory by sweeping the target time $t_\text{tgt}$ across the chunk's frames. At the first overlapping frame $t_k$ between chunk $k$ and chunk $k{+}1$, we take the predicted 3D position $P_i(t_k)$ and apply the already-estimated Sim(3) to express it in chunk $k{+}1$'s local coordinates, yielding the re-query coordinate $\mathbf{p}_i$ for the next chunk. The visual descriptor $S_i$, extracted once from the first frame where the point is visible, is reused across all subsequent chunks without re-extraction. This procedure chains trajectories regardless of whether the point is visible in the overlap: the 3D coordinate is always well-defined, so occluded and out-of-frame points propagate without failure. In contrast, 2D-query methods must project the predicted 3D position back to pixel coordinates to re-query, which is undefined for occluded points and compounds camera prediction error for visible ones. The overall trajectory chaining procedure is described in Algorithm~\ref{alg:traj_chain}.

\subsection{Implementation Details}

We initialize the encoder and geometry heads from Depth Anything 3~\cite{da3} pretrained weights, while the decoder is trained from scratch. We train on a mixture of dynamic datasets (PointOdyssey~\cite{pointodyssey}, Dynamic Replica~\cite{dynamicreplica}, Bedlam2~\cite{bedlam2}, Kubric Movi-F~\cite{kubric}, CoTracker-Kubric~\cite{cotracker3}, Waymo~\cite{waymodrivetrack}, Omniworld~\cite{omniworld}) and static datasets (ScanNet~\cite{scannet}, ScanNet++~\cite{scannet++}, BlendedMVS~\cite{blendedmvs}, Co3Dv2~\cite{co3d}, WildRGBD~\cite{wildrgbd}), where static points are treated as stationary trajectories. Each training sequence contains 16 to 64 frames, with width sampled from $[252,518]$. Further details are in Appendix~\ref{app:a}.

\section{Experiments}
\label{sec:Experiments}
\subsection{Experimental Setting}

\begin{figure}[t]
  \centering
  \includegraphics[width=1.0\linewidth]{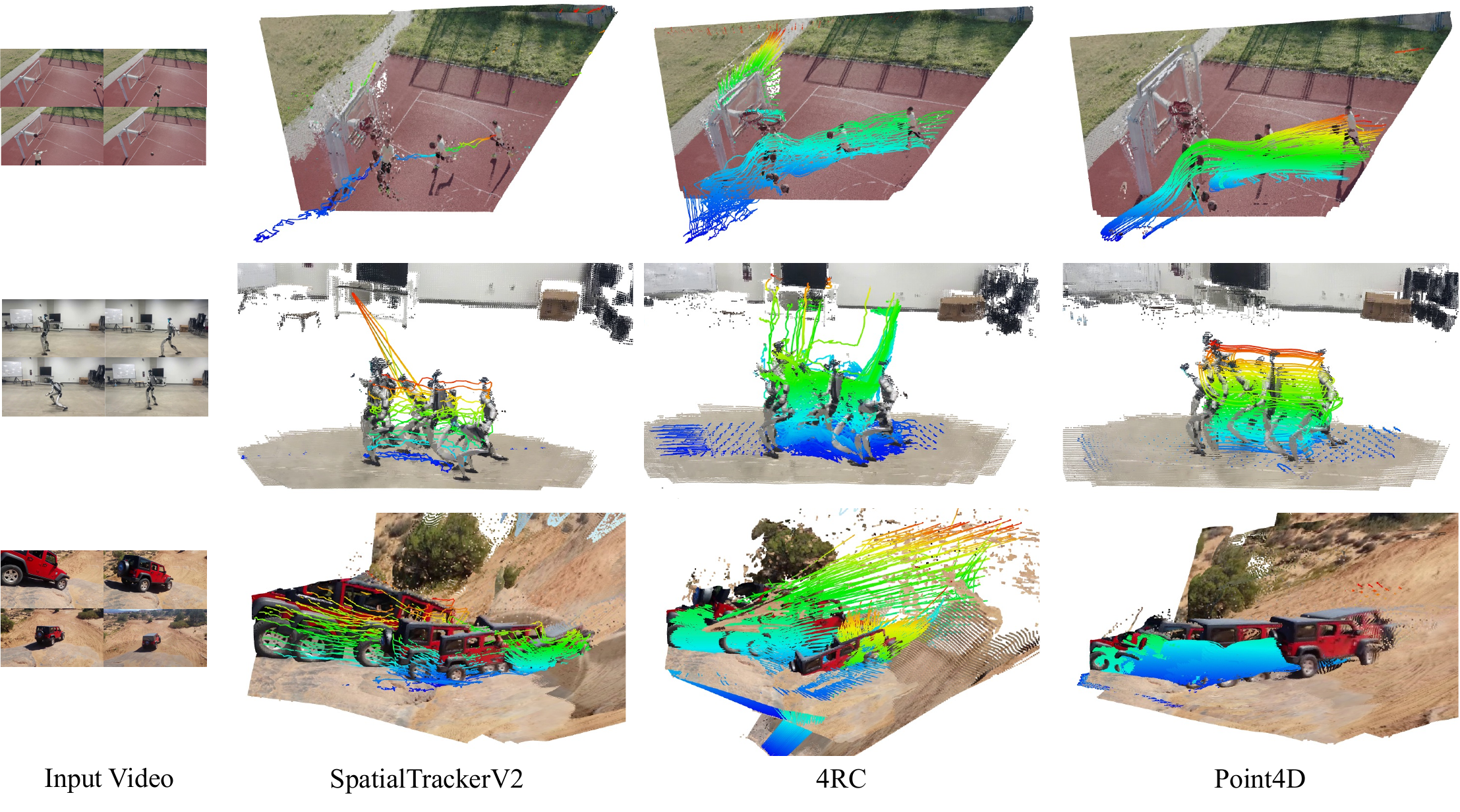}
  \caption{\textbf{Point4D produces reliable and consistent 4D tracking over 200-frame sequences via trajectory chaining}. 4RC struggles to maintain correspondence and its tracks are wrongly chained. SpatialTrackerV2 only can predict a sparser trajectory due to memory constraints.}
  \label{fig:long_chunk}
\end{figure}

\paragraph{Baselines.}
We compare against two categories of methods: (1)~Feed-forward 4D reconstruction methods such as TraceAnything~\cite{traceany}, Any4D~\cite{any4d}, 4RC~\cite{4rc}, and V-DPM~\cite{vdpm}, which predict dense per-pixel 3D positions at queried timesteps and are most directly comparable to ours.
(2)~3D point trackers such as SpatialTrackerV2~\cite{spatrackv2} and TAPIP3D~\cite{tapip3d}, which represent the current state of the art in per-point 3D tracking but do not support dense per-pixel queries and are slower~(Appendix~\ref{app:b}).

\paragraph{Setup.}
We evaluate \ours on (A) long-video tracking, which requires trajectory chaining across chunks, and a simpler (B) single-chunk tracking. For long-video tracking, we use PointOdyssey and Dynamic Replica sequences of 200 frames and TAPVid3D Panoptic Studio~(PStudio) sequences of 150 frames. Each sequence is partitioned into chunks of 48 frames with 8-frame overlap, where, for \ours{}, trajectories are chained across chunks using 3D queries~(Sec.~\ref{sec:3.3.method_chaining}). Feed-forward 4D methods are evaluated with selection-based reprojection chaining, which picks the overlapping frame with the best depth agreement for each point~(Appendix~\ref{app:d}). SpatialTrackerV2 and TAPIP3D use their sliding-window inference modes. For single-chunk tracking, we evaluate on LSFOdyssey, Dynamic Replica and PStudio with up to 64 frames per sequence.

\paragraph{Metrics.}
Following the benchmarking protocol of recent works~\cite{startrack, any4d}, we report end-point error (EPE) and the average percentage of points within distance thresholds $\delta_{3D} \in \{0.1, 0.3, 0.5, 1.0\}$~m~(APD), after median-scale alignment of ground truth and predicted trajectories~(Further details are in Appendix~\ref{app:a}).

\begin{align}
    \text{EPE}_{i,t} &= \|\mathbf{\hat{p}}_{i}^t - \mathbf{p}_{\text{GT},i}^t\| \\
    \text{APD} &= \sum_{i,t} \mathbbm{1} \cdot \left(\text{EPE}_{i,t} < \delta_{\text{3D}}\right)
\end{align}

For long-video, we additionally report Survival rate~\cite{pointodyssey}: the average fraction of video length before tracking failure. A point $i$ is considered failed at frame $t$ if $\left\| \hat{\mathbf{p}}_i^t - \mathbf{p}_{\text{GT},i}^t \right\|_2 > \delta_{\text{3D}}$, and we report:
\begin{equation}
    \text{Survival} = \frac{1}{N} \sum_{i=1}^{N} \frac{t_i^{\text{fail}}-1}{T}
\end{equation}
where $t_i^{\text{fail}}$ is the first failure frame and $N$ is total queries. We average over $\delta_{\text{3D}} \in \{0.1, 0.3, 0.5, 1.0\}$\,m.
\begin{table*}[t]
\centering
\scriptsize
\setlength{\tabcolsep}{6pt}
\setlength{\fboxsep}{1.5pt} 
\renewcommand{\arraystretch}{1.15}
\begin{tabular}{l l ccc ccc ccc}
\toprule
& \multirow{2}{*}{\textbf{Method}} & \multicolumn{3}{c}{\textbf{PointOdyssey}~\cite{pointodyssey}} & \multicolumn{3}{c}{\textbf{Dynamic Replica}~\cite{dynamicreplica}} & \multicolumn{3}{c}{\textbf{PStudio}~\cite{tapvid3d}} \\
\cmidrule(lr){3-5} \cmidrule(lr){6-8} \cmidrule(lr){9-11}
& & EPE $\downarrow$ & APD $\uparrow$ & Survival $\uparrow$ & EPE $\downarrow$ & APD $\uparrow$ & Survival $\uparrow$ & EPE $\downarrow$ & APD $\uparrow$ & Survival $\uparrow$ \\
\midrule
\multirow{2}{*}{\rotatebox[origin=c]{90}{\scriptsize\textbf{Iter.}}}
& TAPIP3D  & 0.952 & 0.417 & 0.317 & \cellcolor{rank2}{0.185} & \cellcolor{rank2}{0.806} & \cellcolor{rank2}{0.748} & \cellcolor{rank1}{0.230} & \cellcolor{rank1}{0.741} & 0.622 \\
& SpatialTrackV2 & \cellcolor{rank1}{0.498} & \cellcolor{rank1}{0.611} & \cellcolor{rank2}{0.477} & \cellcolor{rank3}{0.218} & \cellcolor{rank3}{0.772} & \cellcolor{rank3}{0.687} & \cellcolor{rank2}{0.234} & \cellcolor{rank3}{0.719} & \cellcolor{rank3}{0.623} \\
\midrule
\multirow{5}{*}{\rotatebox[origin=c]{90}{\scriptsize\textbf{Feed-Forward}}}
& TraceAnything  & 2.147 & 0.121 & 0.070 & 0.767 & 0.464 & 0.391 & 0.665 & 0.408 & 0.312 \\
& Any4D  & 1.026 & 0.400 & 0.291 & 0.364 & 0.711 & 0.629 & 0.497 & 0.495 & 0.389 \\
& 4RC   & 0.789 & \cellcolor{rank3}{0.559} & \cellcolor{rank3}{0.463} & 0.336 & 0.733 & 0.654 & 0.379 & 0.613 & 0.534 \\
& VDPM   & \cellcolor{rank3}{0.736} & \cellcolor{rank3}{0.559} & 0.461 & 0.386 & 0.666 & 0.587 & 0.280 & \cellcolor{rank3}{0.719} & \cellcolor{rank2}{0.634} \\
& \textbf{\ours}  & \cellcolor{rank2}{0.616} & \cellcolor{rank2}{0.585} & \cellcolor{rank1}{0.514} & \cellcolor{rank1}{0.155} & \cellcolor{rank1}{0.856} & \cellcolor{rank1}{0.812} & \cellcolor{rank3}{0.236} & \cellcolor{rank2}{0.731} & \cellcolor{rank1}{0.664} \\
\bottomrule
\end{tabular}
\caption{\textbf{Long-Video 4D Tracking via Trajectory Chaining.} \ours{} achieves the best average rank among all compared methods, outperforming every other feed-forward method, while running much faster than iterative trackers (see Appendix). We report EPE~($\downarrow$), APD~($\uparrow$), and survival rate ($\uparrow$) on sequences of 200 frames, partitioned into chunks of 48 frames with 8-frame overlap. \colorbox{rank1}{Red}, \colorbox{rank2}{Orange}, and \colorbox{rank3}{Yellow} indicate the top three results.
\vspace{-0.3in}}
\label{tab:chaining_results}
\end{table*}

\subsection{4D Tracking}
\paragraph{Long-video trajectory chaining.}
Table~\ref{tab:chaining_results} compares all methods on 200-frame sequences (150 for PStudio) that require chaining across multiple chunks. \ours{} outperforms both categories of baselines on most sequences. Feed-forward 4D baselines use reprojection-based chaining, which is undefined for occluded points and compounds camera prediction error even for visible ones, causing trajectories to drift or break at chunk boundaries. Iterative 3D trackers propagate tracks within their own fixed windows, but their iterative refinement is slow and memory constraints preclude dense query sets. In contrast, \ours{} re-queries predicted 3D coordinates directly in each subsequent chunk, avoiding both failure modes.

Figure~\ref{fig:long_chunk} shows qualitative results on 200-frame sequences, where we sample a regular grid of query points on the first frame and decode their trajectories across chunks. \ours{} maintains continuous trajectories through occlusion and field-of-view exits, while 2D-query methods lose track at chunk boundaries. SpatialTrackerV2 uses a sparser query grid due to GPU memory constraints, resulting in visibly sparser trajectories. Additional results can be found in Appendix~\ref{app:e} and \ref{app:f}.

\begin{figure}[t]
  \centering
  \includegraphics[width=\linewidth]{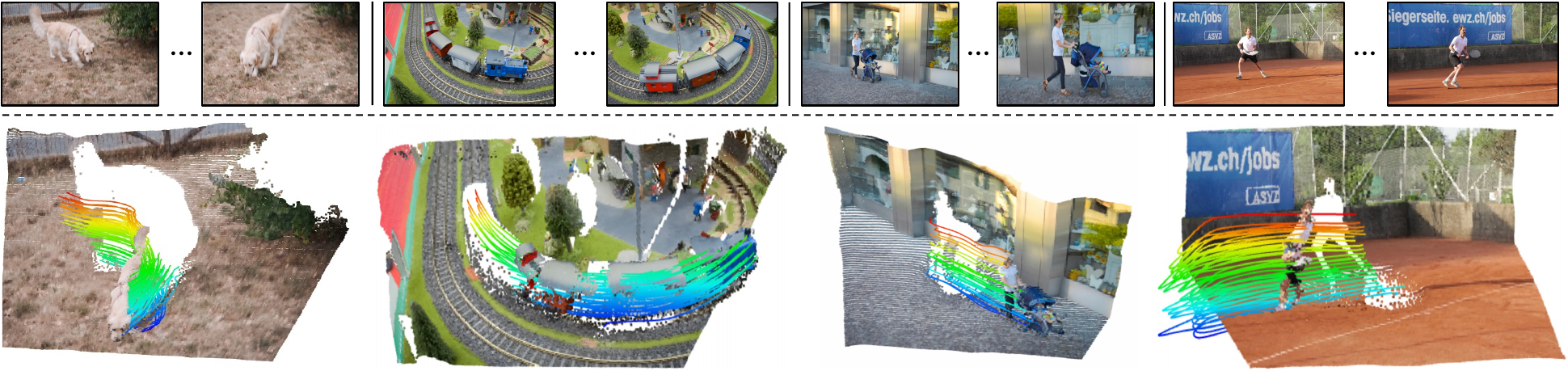}
  \caption{\textbf{Single-chunk 4D tracking on DAVIS~\cite{davis}.} Point4D produces consistent 3D trajectories across challenging real-world sequences.}
  \label{fig:single_chunk}
\end{figure}

\paragraph{Single-chunk tracking.}
Table~\ref{tab:single_chunk_results} evaluates all methods on sequences short enough to fit within a single chunk~(64 frames), isolating decoder accuracy from chaining. \ours performs comparably to other feed-forward 4D methods while outperforming both 2D tracker + 3D reconstruction pipelines and iterative refinement-based methods. Combined with the long-video results, this shows that our advantage on longer sequences comes from reliable 3D-query chaining rather than a gap in single-chunk decoding. Figure~\ref{fig:single_chunk} shows representative results on DAVIS sequences.

\begin{table*}[t]
\centering
\scriptsize
\setlength{\tabcolsep}{6pt}
\renewcommand{\arraystretch}{1.15}
\begin{tabular}{l l cc cc cc}
\toprule
& \multirow{2}{*}{\textbf{Method}} & \multicolumn{2}{c}{\textbf{LSFOdyssey}~\cite{lsfodyssey}} & \multicolumn{2}{c}{\textbf{Dynamic Replica}~\cite{dynamicreplica}} & \multicolumn{2}{c}{\textbf{PStudio}~\cite{tapvid3d}} \\
\cmidrule(lr){3-4} \cmidrule(lr){5-6} \cmidrule(lr){7-8}
& & EPE $\downarrow$ & APD $\uparrow$ & EPE $\downarrow$ & APD $\uparrow$ & EPE $\downarrow$ & APD $\uparrow$ \\
\midrule
\multirow{4}{*}{\rotatebox[origin=c]{90}{\scriptsize\textbf{2D Tracker}}}
& MonST3R + CoTracker3          & 0.61 & 0.51 & 0.81 & 0.43 & 0.51 & 0.52 \\
& MASt3R + CoTracker3          & 0.83 & 0.46 & 0.40 & 0.58 & 0.43 & 0.54 \\
& VGGT + CoTracker3       & 0.47 & 0.59 & 0.26 & 0.69 & 0.26 & 0.69 \\
& MapAnything + CoTracker3          & 0.63 & 0.35 & 0.25 & 0.71 & 0.63 & 0.51 \\
& DepthAnything3 +CoTracker3 & 0.50 & 0.69 & 0.11 & 0.89 & 0.33 & 0.63 \\
\midrule
\multirow{2}{*}{\rotatebox[origin=c]{90}{\scriptsize\textbf{Iter.}}}
& TAPIP3D          & 0.35 & 0.66 & 0.87 & 0.50 & 0.30 & 0.65 \\
& SpatialTrackv2       & 0.34 & 0.68 & 0.69 & 0.62 & \cellcolor{rank2}0.21 & \cellcolor{rank2}0.75 \\
\midrule
\multirow{6}{*}{\rotatebox[origin=c]{90}{\scriptsize\textbf{Feed-Forward}}}
& St4RTrack & 0.56 & 0.48 & 0.17 & 0.81 & 0.41 & 0.53 \\
& TraceAnything   & 0.76 & 0.42 & 0.34 & 0.66 & 0.25 & 0.72 \\
& Any4D            & \cellcolor{rank3}0.27 & 0.72 & \cellcolor{rank1}0.07 & \cellcolor{rank2}0.93 & 0.28 & 0.66 \\
& 4RC              & \cellcolor{rank2}0.16 & \cellcolor{rank1}0.87 & \cellcolor{rank1}0.07 & \cellcolor{rank1}0.95 & 0.29 & 0.66 \\
& VDPM             & \cellcolor{rank1}0.14 & \cellcolor{rank2}0.85 & 0.14 & 0.84 & \cellcolor{rank1}0.17 & \cellcolor{rank1}0.79 \\
& \textbf{\ours}   & \cellcolor{rank3}0.27 & \cellcolor{rank3}0.73 & \cellcolor{rank2}0.09 & \cellcolor{rank3}0.91 & \cellcolor{rank3}0.23 & \cellcolor{rank3}0.73 \\
\bottomrule
\end{tabular}
\caption{\textbf{Single-chunk 4D tracking.} \ours{} performs comparably to
existing methods on short sequences that do not require trajectory
chaining, which confirms that the long-video gains in
Table~\ref{tab:chaining_results} stem from the proposed 3D query formulation,
not a stronger single-chunk decoder. We report EPE ($\downarrow$)
and APD ($\uparrow$). \colorbox{rank1}{Red}, \colorbox{rank2}{Orange},
and \colorbox{rank3}{Yellow} indicate the top three results.}
\label{tab:single_chunk_results}
\end{table*}

\subsection{Ablations and Analysis}

\paragraph{Ablation on query formulation.}
We ablate the two key design choices in Point4D's query formulation: (1) using 3D coordinates instead of 2D pixels, and (2) training with visual descriptors from arbitrary visible frames rather than only the source frame. Table~\ref{tab:query_ablation} compares three variants: \textbf{2D} queries with pixel coordinates $(u,v)$ as in D4RT~\cite{d4rt}; \textbf{3D (source patch)}, which uses 3D coordinates but always extracts the visual descriptor from the source frame, so the model never sees occluded or out-of-frame queries during training; and \textbf{3D} (\ours), which combines 3D coordinates with descriptors from arbitrary visible frames. The 2D variant must reproject each predicted point back to the image plane to continue a trajectory, as in other 2D-query based methods. The 3D source patch variant avoids reprojection but expects a visible query with descriptor from the source frame, where the point may be occluded or out of view at a chunk boundary. \ours{} outperforms both, especially in long-video tracking, indicating that both design choices are needed for reliable chaining.
\begin{table}[t]
\centering
\footnotesize
\setlength{\tabcolsep}{6pt}
\renewcommand{\arraystretch}{1.15}
\begin{tabular}{l cccc cccc}
\toprule
& \multicolumn{4}{c}{\textbf{Long-Video Tracking}} & \multicolumn{4}{c}{\textbf{Single-Chunk Tracking}} \\
\cmidrule(lr){2-5} \cmidrule(lr){6-9}
& \multicolumn{2}{c}{\textbf{PointOdyssey}~\cite{pointodyssey}} & \multicolumn{2}{c}{\textbf{Dynamic Replica}~\cite{dynamicreplica}} & \multicolumn{2}{c}{\textbf{LSFOdyssey}~\cite{lsfodyssey}} & \multicolumn{2}{c}{\textbf{Dynamic Replica}~\cite{dynamicreplica}} \\
\cmidrule(lr){2-3} \cmidrule(lr){4-5} \cmidrule(lr){6-7} \cmidrule(lr){8-9}
\makecell[l]{\textbf{Query}\\\textbf{Formulation}} & EPE $\downarrow$ & SR $\uparrow$ & EPE $\downarrow$ & SR $\uparrow$ & EPE $\downarrow$ & APD $\uparrow$ & EPE $\downarrow$ & APD $\uparrow$ \\
\midrule
2D                  & 0.891 & 0.283 & 0.712 & 0.422 & 0.279 & 0.713 & 0.110 & 0.870 \\
3D (source patch) & 0.869 & 0.380 & 0.825 & 0.266 & 0.442 & 0.620 & 0.121 & 0.800 \\
3D (\ours)                & \textbf{0.616} & \textbf{0.514} & \textbf{0.155} & \textbf{0.812} & \textbf{0.274} & \textbf{0.731} & \textbf{0.091} & \textbf{0.906} \\
\bottomrule
\end{tabular}
\caption{ \textbf{\ours's proposed 3D query formulation is optimal for reliable decoding.} We ablate this choice on both single-chunk and long-video chaining benchmarks. We report EPE~($\downarrow$), APD~($\uparrow$) and Survival Rate~(SR,$\uparrow$).}
\label{tab:query_ablation}
\end{table}

\paragraph{Robustness of tracking across chunks.}

\begin{wrapfigure}{r}{0.5\linewidth}
\vspace{-30pt}
  \centering
  \includegraphics[width=\linewidth]{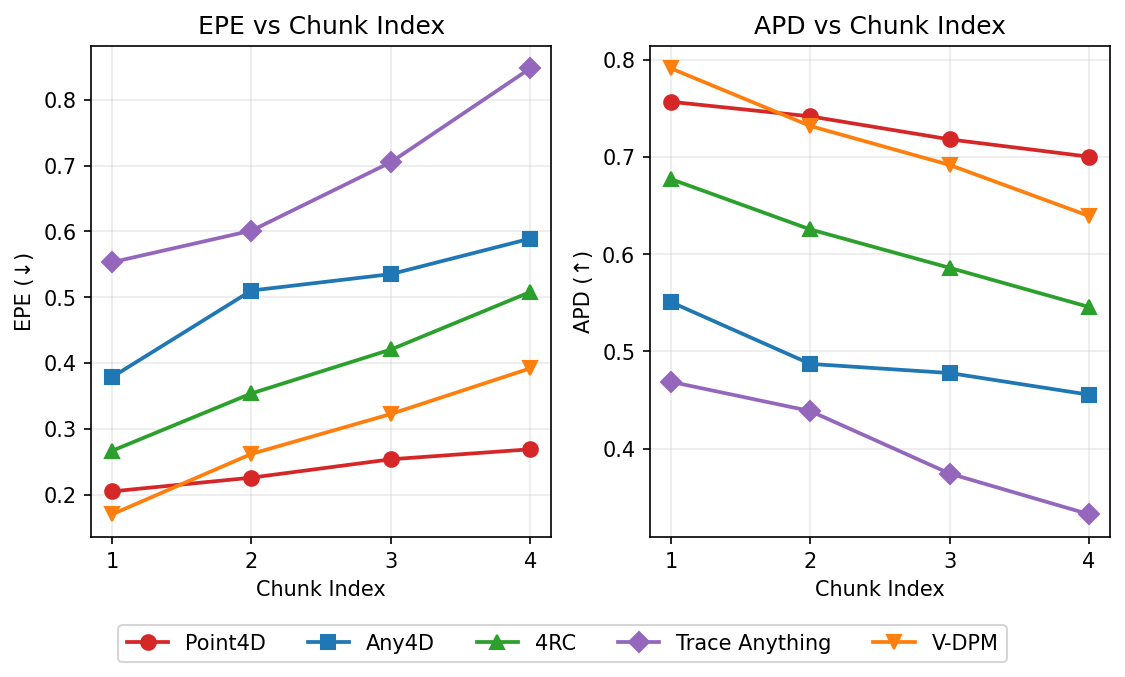}
  \caption{\textbf{Chunk-wise tracking accuracy on PStudio.} \ours~(red) degrades slowly across chunks, while others worsen faster.}
  \label{fig:chunk_metric}
\vspace{-10pt}
\end{wrapfigure}
To verify that our 3D query formulation enables reliable trajectory chaining across video chunks, we measure per-chunk APD and EPE on PStudio long tracking result. Figure~\ref{fig:chunk_metric} shows that \ours{} degrades slowly across chunks in both APD and EPE, while baselines deteriorate faster as chaining proceeds. The gap widens because 2D-query methods compound error at every chunk boundary through reprojection, whereas our 3D re-querying avoids this entirely. Notably, VDPM starts with higher first-chunk accuracy, yet \ours{} surpasses them within a few chunks. This confirms that the advantage of 3D queries lies not in single-chunk decoding but in chaining across many chunks, which matters for long-video 4D reconstruction.

\section{Discussion}
We presented Point4D, a feed-forward model for 4D reconstruction of long-range video sequences. \ours infers motion using 3D coordinate queries compared to prior feed-forward methods which inferred motion from 2D pixels. This decouples trajectory prediction from image-plane visibility, and allows for direct requerying of 3D points across chunks even when occluded or out of field of view. Moreover, we show that our visual descriptor extracted from arbitrary frames improves over the source patch descriptor alone. We show these design choices enable \ours to outperform existing methods on  long-range videos. Ultimately, we believe \ours will serve as a foundation step towards achieving reliable 4D reconstruction on in-the-wild videos of arbitrary length and serve for diverse applications in generative AI, AR/VR and robotics.

\paragraph{Limitations.}
The handoff between chunks carries only each query's 3D coordinate and its patch descriptor; no scene representation or feature memory is retained. Since the encoder represents only what is observed within the current chunk, a point that remains occluded or out of frame for an entire chunk has no supporting evidence in that chunk, and its predicted position becomes unreliable.
Our formulation also relies on predicted depth and on the Sim(3) alignment between consecutive chunks, so errors in depth prediction can compound across chunks.

\paragraph{Acknowledgements.}
We thank the members of the Physical Perception Lab at CMU for their valuable discussions. This work was supported in part by NSF Award IIS-2345610. This work used Bridges-2 at Pittsburgh Supercomputing Center through allocation CIS251064 from the Advanced Cyberinfrastructure Coordination Ecosystem: Services \& Support (ACCESS) program, which is supported by National Science Foundation grants \#2138259, \#2138286, \#2138307, \#2137603, and \#2138296.
This work was supported by Intelligence Advanced Research Projects Activity (IARPA) via Department of Interior/Interior Business Center (DOI/IBC) contract number 140D0423C0074. The U.S. Government is authorized to reproduce and distribute reprints for Governmental purposes notwithstanding any copyright annotation thereon. Disclaimer: The views and conclusions contained herein are those of the authors and should not be interpreted as necessarily representing the official policies or endorsements, either expressed or implied, of IARPA, DOI/IBC, or the U.S. Government.

{
    \small
    \bibliographystyle{ieeenat_fullname}
    \bibliography{main}
}

\newpage
\appendix

\section{Implementation Details}
\label{app:a}
\paragraph{Training Details.}
Table~\ref{tab:training_datasets} summarizes the training datasets and their sampling ratios. OmniWorld~\cite{omniworld} contains dynamic content but lacks trajectory ground truth; we restrict its queries to $t_\text{src} = t_\text{tgt}$, reducing the task to depth and relative-pose estimation without requiring cross-frame trajectory labels.

For each frame, we sample $N=750$ query pixels, with 40\% drawn from edge regions detected by Sobel filtering to encourage coverage of object boundaries. The query indices $t_\text{src}$, $t_\text{tgt}$, and $t_\text{cam}$ are sampled uniformly from the frame indices, with 40\% of queries constrained to $t_\text{cam} = t_\text{tgt}$ so the model frequently predicts points in the target frame's own coordinate system.

We use AdamW with a peak learning rate of $1\times10^{-4}$ for 150 epochs, linearly warmed up till 10 epoch and then cosine-decayed. During training, the learning rate for models initialized from DepthAnything3 is scaled by 0.1. Training is done on 8 H100 GPUs, and the loss for dynamic points is upweighted relative to static points. We apply color jittering, Gaussian blurring, random rescaling, and aspect-ratio augmentation throughout the training.
\begin{table}[h]
\centering
\small
\caption{\textbf{Training datasets.} Sampling ratio denotes the proportion of samples drawn per epoch. Dynamic datasets provide ground-truth trajectories; static datasets treat all points as stationary.}
\label{tab:training_datasets}
\begin{tabular}{lcc}
\toprule
Dataset & Dynamic & Sampling Ratio \\
\midrule
PointOdyssey~\cite{pointodyssey} & \cmark & 19.8\% \\
Dynamic Replica~\cite{dynamicreplica} & \cmark & 19.8\% \\
BEDLAM2~\cite{bedlam2} & \cmark & 19.8\% \\
CoTracker Kubric~\cite{cotracker3} & \cmark & 11.9\% \\
Kubric Movi-F~\cite{kubric} & \cmark & 11.9\% \\
Waymo Drivetrack~\cite{waymodrivetrack} & \cmark & 5.8\% \\
OmniWorld~\cite{omniworld} & \cmark$^\dagger$ & 2.0\% \\
\midrule
ScanNet~\cite{scannet} &\xmark & 2.0\% \\
ScanNet++~\cite{scannet++} & \xmark& 2.0\% \\
BlendedMVS~\cite{blendedmvs} & \xmark& 2.0\% \\
Co3Dv2~\cite{co3d} & \xmark& 2.0\% \\
WildRGBD~\cite{wildrgbd} &\xmark & 1.0\% \\
\bottomrule
\multicolumn{3}{l}{\small $^\dagger$ Dynamic content but no trajectory GT; queries restricted to $t_\text{src} = t_\text{tgt}$.}
\end{tabular}
\end{table}

\paragraph{Evaluation Details.}
For long-video 4D Tracking~(Table~\ref{tab:chaining_results}), we evaluated both static and dynamic points, aligning ground-truth and predicted trajectories with a single global scale per sequence. We include static points because they are the ones that most often leave the field of view during chaining, and thus directly measure tracking accuracy for occluded and out-of-frame points. Results on dynamic points only are reported in Appendix~\ref{app:e}. For single-chunk tracking~(Table~\ref{tab:single_chunk_results}), we evaluate dynamic points only and follow the protocol of Any4D~\cite{any4d}, which rescales each frame pair independently.

\section{Runtime Analysis}
\label{app:b}
We measure the runtime of each method, scaling either the number of input frames or the number of query points at the first frame. When scaling the number of frames~($16$, $32$, $48$, $64$), the number of queries are fixed to $100$, and when scaling the number of queries~($100$ to $5{,}000$) the number of frame is fixed to $48$. All measurements are taken on a single A6000 (48\,GB) at an input resolution of $294\times518$.

\begin{figure}[h]
  \centering
  \includegraphics[width=0.8\linewidth]{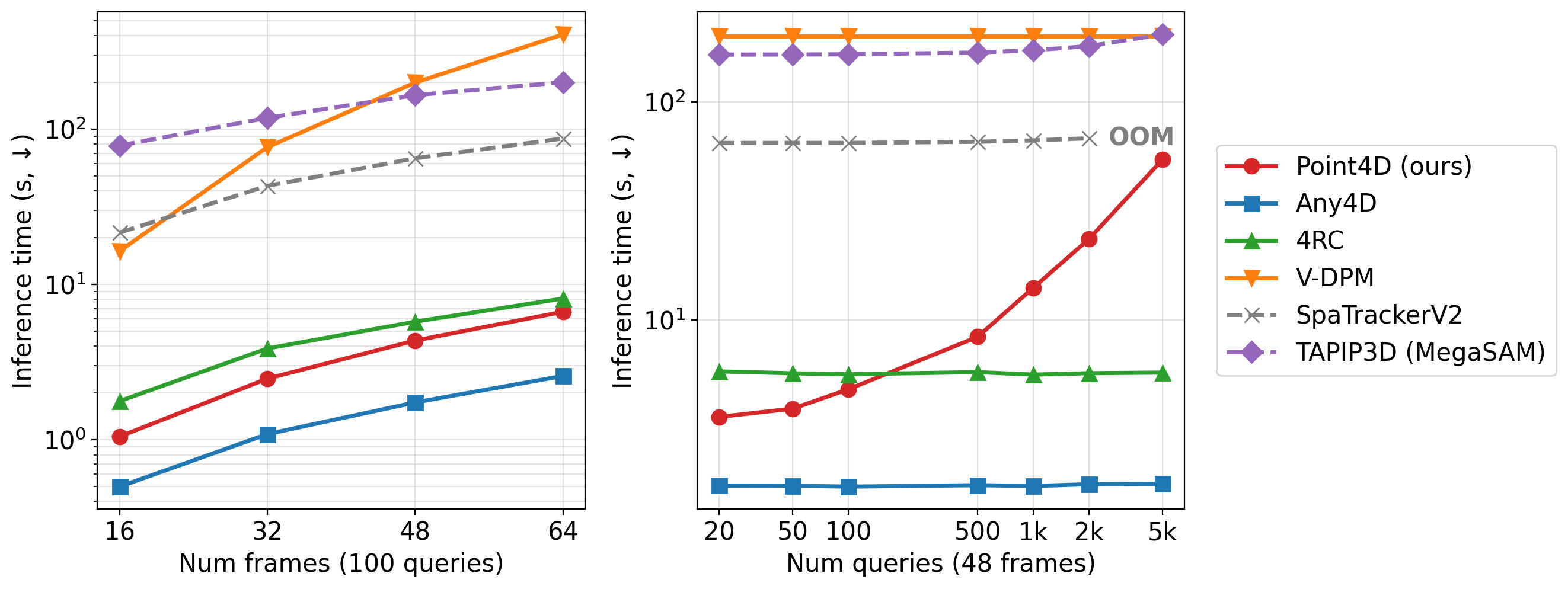}
  \caption{\textbf{Runtime Comparison.} Inference time vs.\ the number of input frames (left) and query points (right). \ours{} is the second fastest with sparse queries; with dense queries it falls behind the dense DPT-head methods but still completes, where SpatialTrackV2 runs out of memory.}
  \label{fig:sup_runtime}
\end{figure}

Results are shown in Figure~\ref{fig:sup_runtime} on a log scale. When scaling the number of input frames, \ours{} is the second-fastest method at every frame count. When scaling the number of queries, the runtime of \ours{} grows, whereas other feed-forward methods remain constant: their DPT-based dense decoders predict every pixel regardless of how many points are queried. This growth is a property of query-based decoding in general, not of 3D queries specifically, and it is precisely what makes it flexible and sparse tracking cheap: querying $100$ points costs proportionally little, while dense decoders pay their full cost either way.
Finally, SpatialTrackerV2~\cite{spatrackv2} and TAPIP3D~\cite{tapip3d} are the slowest methods after VDPM~\cite{vdpm}, and SpatialTrackerV2 runs out of memory at $5{,}000$ queries.

\section{Video depth and camera pose estimation.}
\label{app:c}
We evaluate video depth and camera pose estimation on Sintel~\cite{sintel}, Bonn~\cite{bonn}, and KITTI~\cite{kitti}. Depth accuracy is measured by absolute relative error~(AbsRel, $\downarrow$) and the inlier ratio~$\delta < 1.25$~($\uparrow$). Camera pose is assessed using Absolute Trajectory Error~(ATE), translational Relative Pose Error~(RPE$_t$), and rotational Relative Pose Error~(RPE$_r$), computed after global SE(3) alignment. As shown in Table~\ref{tab:depth_and_pose}, \ours performs on par with DA3, while outperforming other feed-forward 4D reconstruction methods on most metrics. This indicates that \ours{}, built on the DA3 backbone, maintains accurate scene geometry understanding — important for ensuring that the 3D queries used in motion reconstruction are reliably initialized from predicted depth.
\begin{table*}[h]
\centering
\tiny
\setlength{\tabcolsep}{5pt}
\renewcommand{\arraystretch}{1.15}
\begin{tabular}{l cc cc cc ccc ccc}
\toprule
& \multicolumn{6}{c}{\textbf{Video Depth Estimation}} & \multicolumn{6}{c}{\textbf{Camera Pose Estimation}} \\
\cmidrule(lr){2-7} \cmidrule(lr){8-13}
& \multicolumn{2}{c}{\textbf{KITTI}~\cite{kitti}} & \multicolumn{2}{c}{\textbf{Bonn}~\cite{bonn}} & \multicolumn{2}{c}{\textbf{Sintel}~\cite{sintel}} & \multicolumn{3}{c}{\textbf{Bonn}~\cite{bonn}} & \multicolumn{3}{c}{\textbf{Sintel}~\cite{sintel}} \\
\cmidrule(lr){2-3} \cmidrule(lr){4-5} \cmidrule(lr){6-7} \cmidrule(lr){8-10} \cmidrule(lr){11-13}
Method & AbsRel & $\delta\!<\!1.25$ & AbsRel & $\delta\!<\!1.25$ & AbsRel & $\delta\!<\!1.25$ & ATE & RPE$_t$ & RPE$_r$ & ATE & RPE$_t$ & RPE$_r$ \\
\midrule
DA3    & 0.053    & 0.976   & \textbf{0.069}  & \textbf{0.967}    & 0.238  & 0.655    &  0.029   &   0.011 & 0.667 &   0.124  &   0.053  &  0.479 \\
\midrule
TraceAnything & 0.1055 & 0.903 & 6.964 & 0.462 & 0.506 & 0.409 & 0.040 & 0.015 & 43.280 & 0.499 & 0.322 & 10.801\\
Any4D   & 0.090 & 0.939 & 0.414 & 0.600 & 0.670 & 0.384 & 0.059 & 0.023 & 0.632 & 0.619 & 0.241 & 0.741 \\
4RC     & 0.051 & 0.960 & 0.076 & 0.910 & 0.483 & 0.581 & 0.030 & \textbf{0.009} & 0.645 & 0.397 & 0.190 & 0.654 \\
VDPM    & 0.068    & 0.945   & 0.086    & 0.925    & 0.239  & 0.651    & \textbf{0.028}    & \textbf{0.009}    & 0.664    & 0.133    & 0.063    & 0.508    \\
\textbf{\ours}   & \textbf{0.051} & \textbf{0.978} & 0.072 & 0.934 & \textbf{0.202} & \textbf{0.707} & 0.029 & 0.010 & \textbf{0.588} & 0.114 & \textbf{0.049} & \textbf{0.470} \\
\bottomrule
\end{tabular}
\caption{\textbf{Video depth and camera pose estimation.} \ours{} achieves state-of-the-art accuracy among feed-forward 4D reconstruction methods. We report AbsRel ($\downarrow$) and $\delta < 1.25$ ($\uparrow$) for depth, and ATE ($\downarrow$), RPE$_t$ ($\downarrow$), RPE$_r$ ($\downarrow$) for pose on KITTI, Bonn, and Sintel.}
\label{tab:depth_and_pose}
\end{table*}

\section{Chaining Accuracy by Visibility at the Handoff Frame}
\label{app:d}
Chaining requires re-acquiring each point where two chunks meet, and how difficult this is depends on whether the point is visible there. We therefore break down EPE by the visibility state at the handoff frame: \textit{Visible} (the point projects into the frame and is unoccluded), \textit{Occluded} (it projects into the frame but is hidden by another object), and \textit{OOF} (it projects outside the frame boundary). We first identify the strongest handoff strategy for 2D-query baselines, and then compare 2D and 3D re-querying under each condition.

\paragraph{Choosing the handoff strategy for 2D-query baselines.}
\begin{table}[t]
\centering
\caption{\textbf{Chaining strategy ablation for 2D-query chaining (4RC), EPE by visibility.}
Per-sequence EPE averaged over sequences (regime columns average over sequences containing that regime).
Selection-based chaining achieves the lowest EPE on Visible, Occluded, and overall across all datasets.}
\label{tab:chaining_ablation}
\scriptsize
\begin{tabular}{l ccc|c| ccc|c |ccc|c}
\toprule
& \multicolumn{4}{c}{\textbf{PointOdyssey}} & \multicolumn{4}{c}{\textbf{DynamicReplica}} & \multicolumn{4}{c}{\textbf{PStudio}} \\
\cmidrule(lr){2-5}\cmidrule(lr){6-9}\cmidrule(lr){10-13}
\textbf{Chaining} & Vis. & Occ. & OOF & All & Vis. & Occ. & OOF & All & Vis. & Occ. & OOF & All \\
\midrule
First-frame & 0.845 & 1.105 & 1.198 & 0.960 & 0.318 & 0.598 & 0.340 & 0.361 & 0.423 & 0.531 & 0.644 & 0.454 \\
Stationary  & 0.822 & 1.074 & \textbf{0.758} & 0.891 & 0.322 & 0.597 & \textbf{0.260} & 0.357 & 0.421 & 0.525 & \textbf{0.597} & 0.451 \\
Linear      & 0.831 & 1.087 & 0.810 & 0.907 & 0.325 & 0.600 & 0.278 & 0.358 & 0.435 & 0.541 & 0.619 & 0.466 \\
Selection   & \textbf{0.672} & \textbf{0.915} & 1.097 & \textbf{0.789} & \textbf{0.290} & \textbf{0.574} & 0.335 & \textbf{0.336} & \textbf{0.354} & \textbf{0.443} & 0.610 & \textbf{0.379} \\
\bottomrule
\end{tabular}
\end{table}
\begin{table}[t]
\centering
\caption{\textbf{EPE by visibility across datasets.} \ours{} achieves the lowest EPE in nearly every visibility, with the largest margins on Occluded points, where 2D re-querying fails but 3D re-querying carries the point through directly.}
\label{tab:chaining_visibility}
\scriptsize
\begin{tabular}{l ccc|c| ccc|c| ccc|c}
\toprule
& \multicolumn{4}{c}{\textbf{PointOdyssey}} & \multicolumn{4}{c}{\textbf{DynamicReplica}} & \multicolumn{4}{c}{\textbf{PStudio}} \\
\cmidrule(lr){2-5}\cmidrule(lr){6-9}\cmidrule(lr){10-13}
\textbf{Method} & Vis. & Occ. & OOF & All & Vis. & Occ. & OOF & All & Vis. & Occ. & OOF & All \\
\midrule
TraceAnything & 2.049 & 2.258 & 2.748 & 2.147 & 0.728 & 0.856 & 1.203 & 0.767 & 0.635 & 0.744 & 1.187 & 0.665 \\
Any4D  & 0.893 & 1.190 & 1.377 & 1.026 & 0.305 & 0.635 & 0.509 & 0.364 & 0.474 & 0.582 & 0.802 & 0.497 \\
4RC    & 0.672 & 0.915 & \textbf{1.097} & 0.789 & 0.290 & 0.574 & 0.335 & 0.336 & 0.354 & 0.443 & 0.610 & 0.379 \\
VDPM  & 0.636 & 0.860 & 1.171 & 0.736 & 0.346 & 0.607 & 0.379 & 0.386 & 0.261 & 0.332 & 0.493 & 0.280 \\
\ours  & \textbf{0.548} & \textbf{0.597} & 1.147 & \textbf{0.616} & \textbf{0.140} & \textbf{0.202} & \textbf{0.309} & \textbf{0.155} & \textbf{0.228} & \textbf{0.265} & \textbf{0.409} & \textbf{0.236} \\
\bottomrule
\end{tabular}
\end{table}
Since 2D-query methods require projecting predicted 3D points to pixel coordinates for re-querying, the choice of handoff strategy affects chaining quality. Table~\ref{tab:chaining_ablation} ablates four strategies using 4RC as the base method across three datasets. Each strategy decides, from the trajectory predicted in the current chunk, where to re-query the point in the next chunk. \textit{First-frame} projects all points to the first overlapping frame and clips out-of-frame projections to the image boundary. \textit{Stationary} and \textit{Linear} apply the same projection but extrapolate out-of-frame points as stationary or with constant velocity, respectively. \textit{Selection} projects each point at every overlapping frame, selects the frame where the projected depth best matches the depth head's prediction, and re-queries at that frame, which reduces occlusion-related errors by choosing the frame where the point is most likely visible. \textit{Selection} attains the lowest EPE overall on all three datasets, and we therefore use it for all 2D-query baselines.

\paragraph{3D vs.\ 2D re-querying across visibility regimes.}
Table~\ref{tab:chaining_visibility} compares \ours{} with 2D-query baselines (all using Selection chaining) broken down by visibility regime. \ours{} achieves lower EPE on visible and occluded points on every dataset, with the largest margin on occluded points. This indicates that even when a 2D method re-queries at the best overlapping frame they still cannot successfully chain occluded points, whereas a 3D query carries it through the occlusion directly. For out-of-frame points the advantage narrows, and on PointOdyssey 4RC is better. This is expected: some OOF points the next chunk never observes leave no trace in its scene representation, while 2D methods clip to the image boundary and re-query a wrong surface that can still yield a smaller error.

\vspace{-0.15in}

\section{Additional Quantitative Results on Long-Video Tracking}
\label{app:e}
\paragraph{4D tracking results on dynamic points.}
\begin{table*}[t]
\centering
\scriptsize
\setlength{\tabcolsep}{6pt}
\setlength{\fboxsep}{1.5pt}
\renewcommand{\arraystretch}{1.15}
\begin{tabular}{l l ccc ccc}
\toprule
& \multirow{2}{*}{\textbf{Method}} & \multicolumn{3}{c}{\textbf{PointOdyssey}~\cite{pointodyssey}} & \multicolumn{3}{c}{\textbf{Dynamic Replica}~\cite{dynamicreplica}} \\
\cmidrule(lr){3-5} \cmidrule(lr){6-8}
& & EPE $\downarrow$ & APD $\uparrow$ & Survival $\uparrow$ & EPE $\downarrow$ & APD $\uparrow$ & Survival $\uparrow$ \\
\midrule
\multirow{2}{*}{\rotatebox[origin=c]{90}{\scriptsize\textbf{Iter.}}}
& TAPIP3D  & \cellcolor{rank3}{0.779} & \cellcolor{rank3}{0.470} & 0.353 & \cellcolor{rank3}{0.251} & \cellcolor{rank3}{0.728} & \cellcolor{rank3}{0.614} \\
& SpatialTrackV2 & \cellcolor{rank1}{0.493} & \cellcolor{rank1}{0.589} & \cellcolor{rank2}{0.425} & \cellcolor{rank2}{0.240} & \cellcolor{rank2}{0.755} & \cellcolor{rank2}{0.636} \\
\midrule
\multirow{5}{*}{\rotatebox[origin=c]{90}{\scriptsize\textbf{Feed-Forward}}}
& TraceAnything  & 1.863 & 0.120 & 0.070 & 0.722 & 0.462 & 0.361 \\
& Any4D  & 1.127 & 0.276 & 0.170 & 0.431 & 0.614 & 0.505 \\
& 4RC   & 0.879 & 0.445 & 0.355 & 0.359 & 0.698 & 0.598 \\
& VDPM   & 0.784 & 0.467 & \cellcolor{rank3}{0.373} & 0.386 & 0.660 & 0.557 \\
& \textbf{\ours}  & \cellcolor{rank2}{0.617} & \cellcolor{rank2}{0.566} & \cellcolor{rank1}{0.474} & \cellcolor{rank1}{0.200} & \cellcolor{rank1}{0.794} & \cellcolor{rank1}{0.717} \\
\bottomrule
\end{tabular}
\caption{\textbf{Long-Video 4D Tracking on Dynamic Points Only.} \ours{} achieves the best
average rank among all compared methods, outperforming every other feed-forward methods. We report metrics over dynamic points only, on sequences of 200 frames partitioned into chunks of 48 frames with 8-frame overlap. PStudio is omitted as it evaluates only dynamic queries, making the results identical to Table~\ref{tab:chaining_results}. \colorbox{rank1}{Red}, \colorbox{rank2}{Orange}, and \colorbox{rank3}{Yellow} indicate the top three results.}
\label{tab:chaining_results_dynamic}
\end{table*}
Table~\ref{tab:chaining_results} evaluates all query points, including those on static background, whose trajectories are largely explained by camera motion alone. To verify that \ours{} remains effective on points that actually move, we repeat the evaluation using only dynamic points. PStudio is excluded, as its ground-truth trajectories are annotated only for dynamic points and its numbers are therefore identical. As shown in Table~\ref{tab:chaining_results_dynamic}, \ours{} outperforms every feed-forward method on all metrics and attains the best average rank among all compared methods, showing that 3D-query chaining extends trajectories reliably even when the queried points are in motion.

\paragraph{4D tracking results on longer sequences.}
\begin{table*}[t]
\centering
\scriptsize
\setlength{\tabcolsep}{6pt}
\setlength{\fboxsep}{1.5pt}
\renewcommand{\arraystretch}{1.15}
\begin{tabular}{l l ccc ccc}
\toprule
& \multirow{2}{*}{\textbf{Method}} & \multicolumn{3}{c}{\textbf{PointOdyssey}~\cite{pointodyssey} (500 frames)} & \multicolumn{3}{c}{\textbf{Dynamic Replica}~\cite{dynamicreplica} (300 frames)} \\
\cmidrule(lr){3-5} \cmidrule(lr){6-8}
& & EPE $\downarrow$ & APD $\uparrow$ & Survival $\uparrow$ & EPE $\downarrow$ & APD $\uparrow$ & Survival $\uparrow$ \\
\midrule
\multirow{2}{*}{\rotatebox[origin=c]{90}{\scriptsize\textbf{Iter.}}}
& TAPIP3D  & \cellcolor{rank3}{1.020} & \cellcolor{rank3}{0.421} & \cellcolor{rank3}{0.292} & \cellcolor{rank2}{0.198} & \cellcolor{rank2}{0.797} & \cellcolor{rank2}{0.725} \\
& SpatialTrackV2 & \cellcolor{rank2}{1.008} & \cellcolor{rank2}{0.461} & \cellcolor{rank2}{0.307} & \cellcolor{rank3}{0.251} & \cellcolor{rank3}{0.749} & \cellcolor{rank3}{0.644} \\
\midrule
\multirow{3}{*}{\rotatebox[origin=c]{90}{\scriptsize\textbf{F.F.}}}
& 4RC   & 1.597 & 0.338 & 0.228 & 0.418 & 0.676 & 0.582 \\
& VDPM   & 1.458 & 0.318 & 0.213 & 0.478 & 0.610 & 0.513 \\
& \textbf{\ours}  & \cellcolor{rank1}{0.972} & \cellcolor{rank1}{0.482} & \cellcolor{rank1}{0.387} & \cellcolor{rank1}{0.174} & \cellcolor{rank1}{0.836} & \cellcolor{rank1}{0.786} \\
\bottomrule
\end{tabular}
\caption{\textbf{Long-Video 4D Tracking on Extended Sequences.} \ours{} ranks first on
every metric, surpassing not only feed-forward baselines but also iterative trackers. Chunking follows the
same setting (48-frame chunks with 8-frame overlap), so longer sequences require more
handoffs: $12$ for PointOdyssey and $7$ for Dynamic Replica. \colorbox{rank1}{Red}, \colorbox{rank2}{Orange},
and \colorbox{rank3}{Yellow} indicate the top three results.}
\label{tab:chaining_results_longer}
\end{table*}
To assess how each method behaves under more chaining steps, we extend the evaluation to 500 frames on PointOdyssey and 300 frames on Dynamic Replica (the full length of its sequences), keeping the chunk size 48 and overlap 8; PStudio is excluded as it is already evaluated at its full length of 150 frames. The results are shown in Table~\ref{tab:chaining_results_longer}. \ours{} ranks first on every metric, surpasses all iterative trackers and feed-forward 4D models at longer horizons. This demonstrates the robustness of our method for long video tracking.

\begin{figure}[h]
  \centering
  \includegraphics[width=1.0\linewidth]{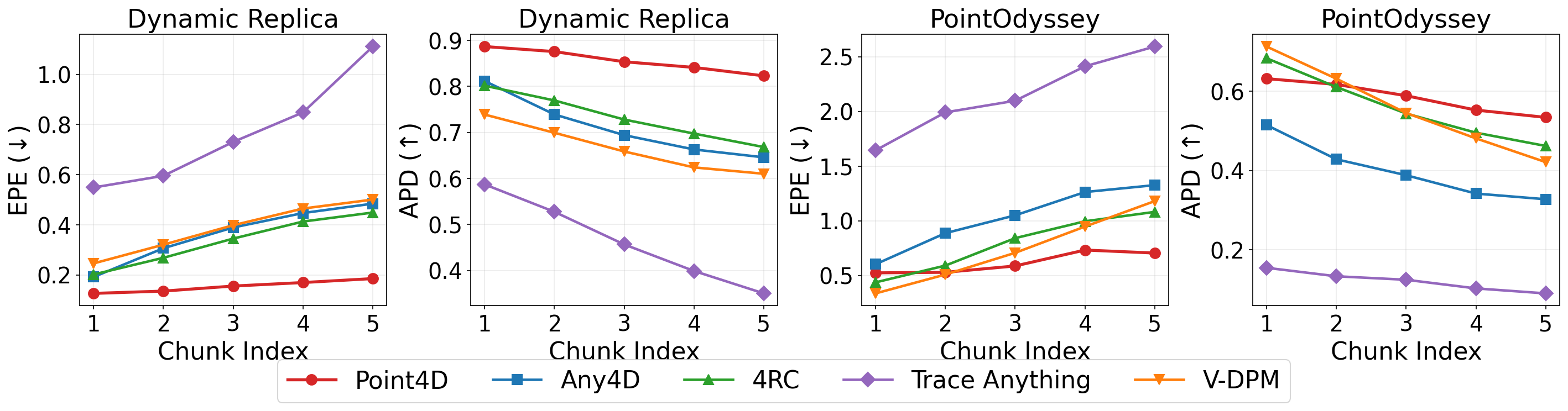}
  \caption{\textbf{Chunk-wise tracking accuracy on (Left)~Dynamic Replica and (Right)~PointOdyssey.} On both datasets, the accuracy of \ours{}~(red) degrades more slowly than that of other methods as chaining proceeds, demonstrating the robustness of 3D-query trajectory chaining.}
  \label{fig:sup_chunk_metric}
\end{figure}

\paragraph{Additional chunk-wise tracking accuracy analysis.}
Complementing the chunk-wise robustness analysis in Figure~\ref{fig:chunk_metric}, we report the full results on PointOdyssey and Dynamic Replica in Figure~\ref{fig:sup_chunk_metric}. On both datasets, the accuracy of \ours{} remains stable across chunks, while the accuracy of feed-forward methods which rely on 2D re-querying degrades as chaining proceeds.

\section{Additional Qualitative Results on Long-Video Tracking}
\label{app:f}
\begin{figure}[t]
  \centering
  \includegraphics[width=\linewidth]{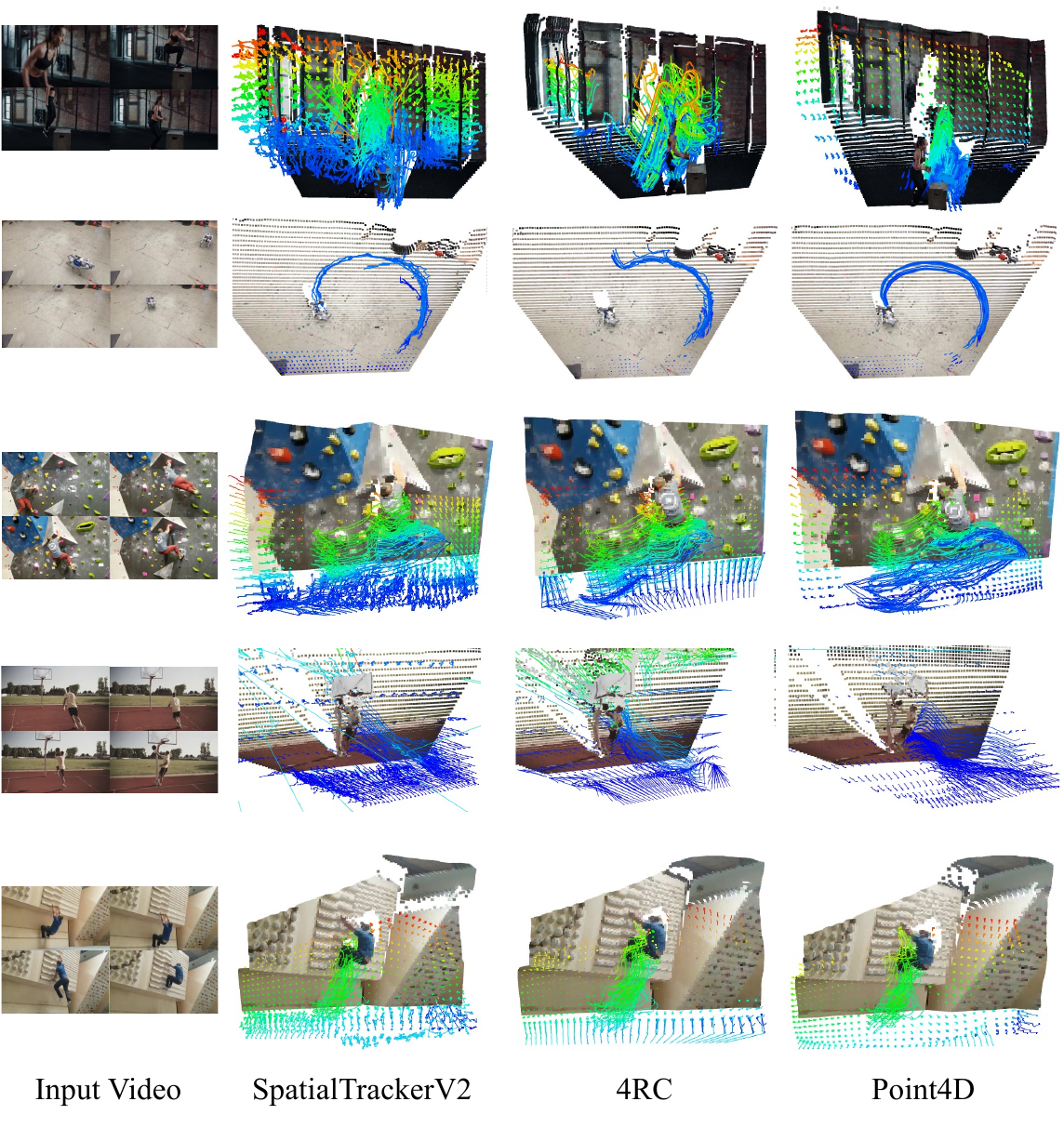}
  \caption{\textbf{Additional results for long-video tracking via trajectory chaining.} Point4D produces consistent 3D trajectories across long videos.}
  \label{fig:sup_result}
\end{figure}

\begin{figure}[t]
  \centering
  \includegraphics[width=\linewidth]{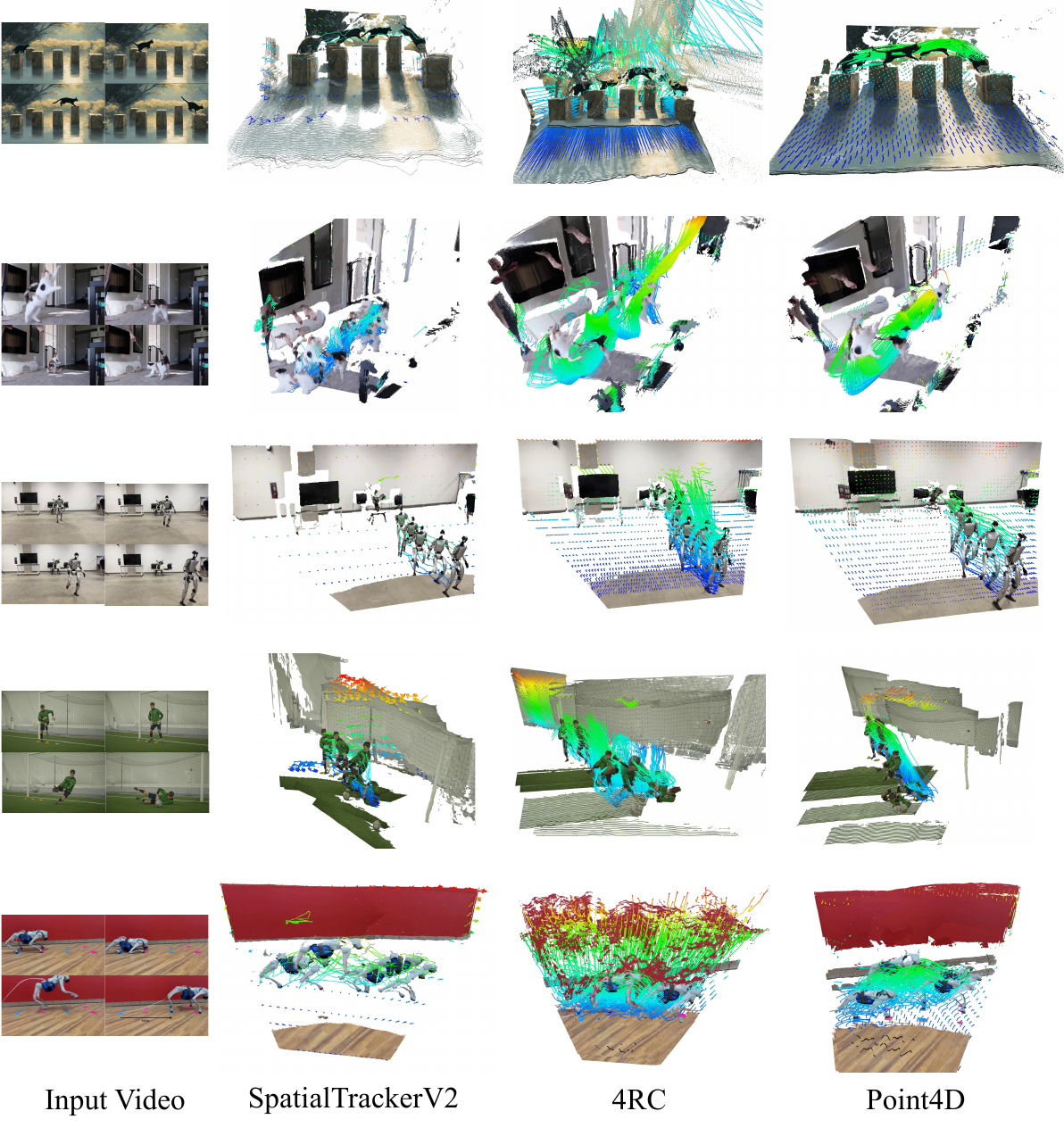}
  \caption{\textbf{Additional results for long-video tracking via trajectory chaining.} Point4D produces consistent 3D trajectories across long videos.}
  \label{fig:sup_result_2}
\end{figure}

We provide additional tracking results for long videos, where the number of frames varies between 100 and 500 in Figure~\ref{fig:sup_result} and Figure~\ref{fig:sup_result_2}. Each chunk has size 48 with 16 overlapping frames. \ours{} show robust trajectory prediction across long videos, showing its effectiveness on trajectory chaining.

\vspace{0.1in}

\clearpage

\clearpage


\end{document}